\documentclass[10pt,twocolumn,letterpaper]{article}

\usepackage[pagenumbers]{cvpr} 

\usepackage{graphicx}
\usepackage{amsmath}
\usepackage{amssymb}
\usepackage{booktabs}
\usepackage[accsupp]{axessibility}
\usepackage[colorlinks,linkcolor=red]{hyperref}

\usepackage{tabularray}
\usepackage{adjustbox}
\usepackage{booktabs}

\usepackage[capitalize]{cleveref}
\crefname{section}{Sec.}{Secs.}
\Crefname{section}{Section}{Sections}
\Crefname{table}{Table}{Tables}
\crefname{table}{Tab.}{Tabs.}

\def\cvprPaperID{7049} 
\def\confName{CVPR}
\def\confYear{2023}

\newcommand\Name{GeoMAE}

\begin{document}

\title{GeoMAE: Masked Geometric Target Prediction \\for Self-supervised Point Cloud Pre-Training}


\author{Xiaoyu Tian$\phantom{}^{1}$\hspace{15pt}
Haoxi Ran$\phantom{}^{2}$\hspace{15pt}
Yue Wang$\phantom{}^{3}$\hspace{15pt}
Hang Zhao$\phantom{}^{1}$\thanks{Corresponding to: hangzhao@mail.tsinghua.edu.cn} \vspace{10pt} \\
$\phantom{}^1$IIIS, Tsinghua University\hspace{10pt}
$\phantom{}^2$CMU\hspace{10pt}
$\phantom{}^3$NVIDIA\hspace{10pt} 
}

\maketitle

\begin{abstract}
This paper tries to address a fundamental question in point cloud self-supervised learning: what is a good signal we should leverage to learn features from point clouds without annotations? To answer that, we introduce a point cloud representation learning framework, based on geometric feature reconstruction. In contrast to recent papers that directly adopt masked autoencoder (MAE) and only predict original coordinates or occupancy from masked point clouds, our method revisits differences between images and point clouds and identifies three self-supervised learning objectives peculiar to point clouds, namely centroid prediction, normal estimation, and curvature prediction. Combined with occupancy prediction, these four objectives yield a nontrivial self-supervised learning task and mutually facilitate models to better reason fine-grained geometry of point clouds.
Our pipeline is conceptually simple and it consists of two major steps: first, it randomly masks out groups of points, followed by a Transformer-based point cloud encoder; second, a lightweight Transformer decoder predicts centroid, normal, and curvature for points in each voxel. We transfer the pre-trained Transformer encoder to a downstream peception model. On the nuScene Datset, our model achieves 3.38 mAP improvement for object detection, 2.1 mIoU gain for segmentation, and  1.7 AMOTA gain for multi-object tracking. We also conduct experiments on the Waymo Open Dataset and achieve significant performance improvements over baselines as well.
\footnote{Our code is available at \url{https://github.com/Tsinghua-MARS-Lab/GeoMAE}.}

\end{abstract}

\section{Introduction}
\begin{figure}[ht!]
  \centering

  \includegraphics[width=\columnwidth]{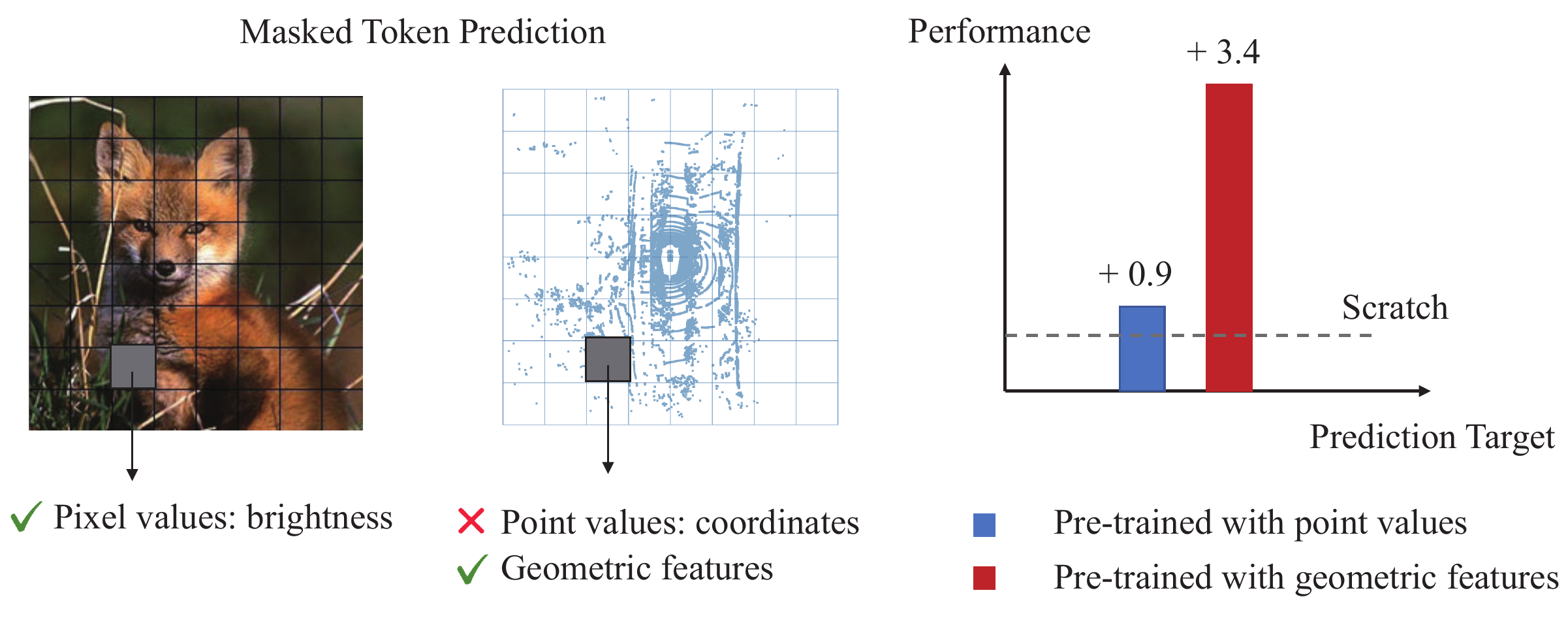}
    \caption{Pixel value regression has been proved effective in masked autoencoder pre-training for images. We find this practice ineffective in point cloud pre-training and propose a set of geometry aware prediction targets.}
   \label{fig:difference}
   \vspace{-8pt}
\end{figure}

While object detection and segmentation from LiDAR point clouds have achieved significant progress, these models usually demand a large amount of 3D annotations that are hard to acquire. To alleviate this issue, recent works explore learning representations from unlabeled point clouds, such as contrastive learning~\cite{pointcontrast,depthcontrast,STRL}, and mask modeling~\cite{Point-bert,point-mae}. Similar to image-based representation learning settings, these representations are transferred to downstream tasks for weight initialization. However, the existing self-supervised pretext tasks do not bring adequate improvements to the downstream tasks as expected. 

Contrastive learning based methods typically encode different `views' (potentially with data augmentation) of point clouds into feature space. They bring features of the same point cloud closer and make features of different point clouds `repel' each other. Other recent works use masked modeling to learn point cloud features through self reconstruction~\cite{Point-bert,point-mae}. That is, randomly sparsified point clouds are encoded by point cloud feature extractors, followed by a reconstruction module to predict original point clouds. These methods, when applied to point clouds, ignore the fundamental difference of point clouds from images -- point clouds provide scene geometry while images provide brightness. As shown in Figure~\ref{fig:difference}, this modality disparity hampers direct use of methods developed in the image domain for point cloud domain, and thus calls for novel self-supervised objectives dedicated to point clouds. 

Inspired by modeling and computational techniques in geometry processing, we introduce a self-supervised learning framework dedicated to point clouds. Most importantly, we design a series of prediction targets which describe the fine-grained geometric features of the point clouds. These geometric feature prediction tasks jointly drive models to recognize different shapes and areas of scenes. Concretely, our method starts with a point cloud voxelizer, followed by a feature encoder to transform each voxel into a feature token. These feature tokens are randomly dropped based on a pre-defined mask ratio. Similar to the original MAE work~\cite{mae}, visible tokens are encoded by a Transformer encoder. Then a Transformer decoder reconstructs the features of the original voxelized point clouds. Finally, our model predicts point statistics and surface properties in parallel branches. 

We conduct experiments on a diverse set of outdoor point cloud datasets including nuScenes~\cite{nuscenes} and Waymo~\cite{waymo}. Our setting consists of a self-supervised pre-training stage and a downstream task stage (3D detection, 3D tracking, segmentation), where they share the same point cloud backbone. Our results show that even without additional unlabeled point clouds, self-supervised pre-training with objectives proposed by this paper can significantly boost the performance of 3D object detection. To summarize, our contributions are: 
\begin{itemize}
    \item We introduce geometry aware self-supervised objectives for point clouds pre-training. Our method leverages fine-grained point statistics and surface properties to enable effective representation learning.
    \item With our novel learning objectives, we achieve state-of-the-art performance compared to previous 3D self-supervised learning methods on a variety of downstream tasks including 3D object detection, 3D/BEV segmentation, and 3D multi-object tracking.
    \item We conduct comprehensive ablation studies to understand the effectiveness of each module and learning objective in our approach.
\end{itemize}

\section{Related Work}
\begin{figure*}[ht!]
  \centering
   \includegraphics[scale=0.6]{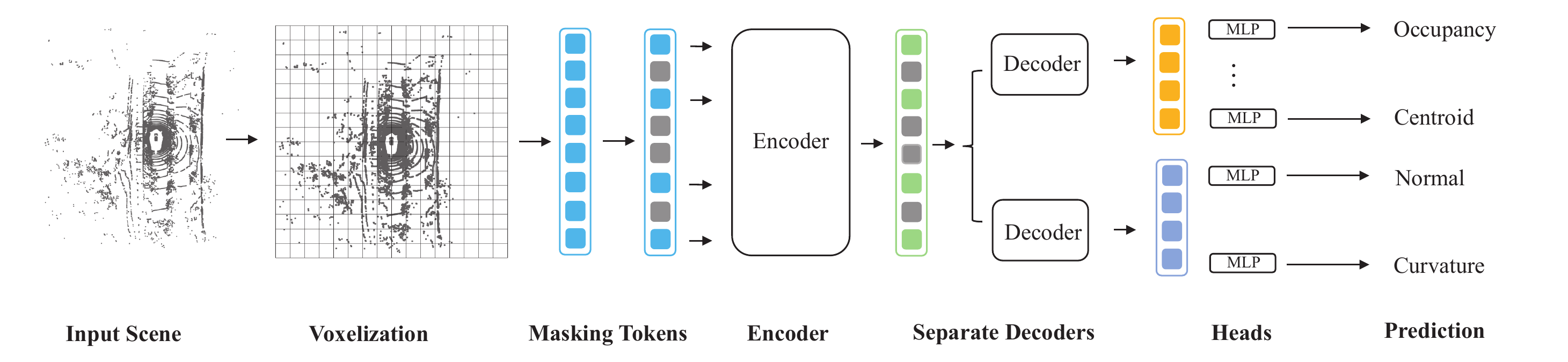}
   \caption{Architecture Overview. The input point cloud scene is first voxelized into voxel grids. After the voxelization, we randomly mask the voxel tokens and fed the visible ones into a sparse encoder-decoder transformer. The encoder is encouraged to capture the geometric information of the point cloud by supervising our proposed geometric prediction targets.}
   \label{fig:pipeline}
\end{figure*}

\subsection{Self-Supervised Learning for Point Clouds} 
Self-supervised learning for point cloud \cite{pointcontrast,depthcontrast,STRL,3dssl-partseg,3dssl-gan,3dssl-global-local,sonet,occo,Point-bert,point-mae} has drawn considerable attention due to the expensive cost of labeling the 3D point cloud. Some are based on contrastive paradigms~\cite{pointcontrast,depthcontrast,STRL}. PointContrast~\cite{pointcontrast} learns from correspondences between different point cloud views with a contrastive loss. DepthContrast~\cite{depthcontrast} considers different depth map as an instance and discriminating between them to learn the representation. STRL~\cite{STRL} learns the invariant representation from two augmented temporally-correlated frames from a 3D point cloud sequence. Others \cite{3dssl-partseg,3dssl-gan,3dssl-global-local} utilize a pretext task to promote self-supervised representation learning.
\cite{3dssl-partseg} phrases the pretext task as a part segmentation task by displacing the part of the parts of the point cloud and then predicting their ordering labels. \cite{3dssl-gan} squeezes learned representations through an implicitly defined parametric discrete generative model bottleneck. \cite{3dssl-global-local} introduces a bidirectional reasoning between local and global to capture the underlying semantic knowledge. Motivated by the huge success of 2D masked image modeling, masked point modeling methods \cite{Point-bert,point-mae} have been proposed recently. Point-BERT \cite{Point-bert} adopts a BERT-style pre-training strategy by predicting discrete tokens of masked input point parts. Point-MAE \cite{point-mae} simply predicts the original coordinates of the masked point patches tokens.

\subsection{Masked Image Modeling} 
Motivated by the success of BERT \cite{bert} for masked language modeling, Masked Image Modeling (MIM)~\cite{beit,mae,simmim,maskfeat,peco,cae,a2mim,data2vec,ibot} becomes a popular pretext task for self-supervised visual representation learning. BEiT \cite{beit} first introduces the pre-training pattern of BERT into the computer vision field by masking out the random image patches and predicting discrete tokens. MAE \cite{mae} and SimMIM \cite{simmim} both propose to predict the raw pixels of the masked patches. Compared with SimMIM, MAE is more pre-training efficient because it only takes the visible token as the input of the encoder and passes all tokens through a lightweight decoder. Many following works use such asymmetric architecture but explore different prediction targets. MaskFeat \cite{maskfeat} uses low-level local features HOG \cite{hog} as the prediction target. $A^{2}$MIM \cite{a2mim} introduce to learn the frequency component of the masked patch features. PeCo \cite{peco} uses an offline visual perceptual codebook to guide the training. 

\subsection{Geometry Learning in Point Cloud}
In computer graphics, previous works \cite{taubin,zhang2008curvature,hoppe1992surface,welch1994free,mitra2003estimating,pointshop} propose various methods for the calculation of differential properties of 3D discrete geometry. Curvature and normal are two of these most important properties. Taubin algorithm \cite{taubin} proposes to estimate the curvature of a surface at each point of a polyhedral approximation. CAN \cite{zhang2008curvature} introduces Local Fitting for normal curvatures by employing chord, neighbor normal vector and osculating circle. As for surface normal estimation, Hoppe \etal \cite{hoppe1992surface} first suggests to fit a least square plane to k nearest neighbors of each point to estimate its normal. Mitra \etal \cite{mitra2003estimating} analyzes the methods of least square with noise added and provides theoretical bound.
\par
In deep learning field, methods\cite{pointnet,pointnet++,dgcnn,kpconv,Minkowski,voxelnet,voxnet} are commonly based
on some assumptions of implicit local geometry. Point-based methods \cite{pointnet,pointnet++,pointcnn,dgcnn,point2sequence,repsurf,kpconv} usually adopt set abstraction to capture local points features region-wise. Voxel-based methods \cite{Minkowski,voxelnet,dynamic-voxelization,pointpillars} project the point clouds to 3D voxel grids and encode features of points inside the same voxel by voxelization.

\section{Method}

\subsection{Architecture Overview}
We propose a simple yet effective method for self-supervised point cloud representation learning, named \textbf{\Name}. \Name predicts both point statistics and surface geometric properties from point clouds. The overall pipeline is illustrated in Figure~\ref{fig:pipeline}. First, we voxelize the original point cloud and transform it into voxel patch tokens. We randomly masked out voxel tokens based on pre-defined ratios for a challenging pre-text task. We define a set of learnable tokens for masked tokens. These visible tokens (corresponding to masked tokens) are fed into a sparse transformer encoder. Conditioned on features of visible tokens, learnable masked tokens are processed by separated decoders to predict both point statistics (centroid and occupancy) and surface properties (normal and curvature). Next, we will elaborate on each step. 

\paragraph{Voxel Token Embedding and Masking.} We follow recent 3D perception architectures~\cite{second} and transform sparse input point clouds into regular voxel grids. Then, these voxels are processed by 3D convolutional neural networks or transformer-based networks. We adopt the widely-used dynamic voxelization \cite{dynamic-voxelization} to perform voxelization: First, the input scene is divided into equally spaced voxels as shown in Figure~\ref{fig:pipeline}. Each point ${p_{i}}$ will be assigned to a voxel ${v_{j}}$ where the point resides. Then, we pass non-empty voxels through VFE \cite{voxelnet} layers to obtain per-voxel features/tokens ${T_{v}}$.  
Based on evidence from 2D masked modeling methods~\cite{mae}, we choose a high mask ratio ($70\%$) when removing tokens. Our method predicts target properties per learnable masked token.

\paragraph{Sparse Encoder.} After random masking, only visible voxel tokens are fed into an encoder. Due to the sparse and long-range nature of the input scene, we choose a sparse transformer proposed in SST \cite{sst} as our encoder. Similar to Swin-Transformer \cite{swin-transformer}, self-attention is only calculated among non-empty voxels within the same region in SST. 
The output token of the encoder is $T_e$, together with the learnable masked token $T_m$ to form the input $T_d$ of the decoder.

\paragraph{Decoders.} We use two separate decoders to decode point statistics and surface properties, respectively.
Each decoder consists of two sparse transformer blocks. These two decoders take the same input ${T_\text{d}}$ and generate two output features ${T_\text{point}}$ and ${T_\text{surface}}$. Empirically, we found such a separate design better facilitated models to learn point statistics and surface properties than a single shared decoder did. Finally, we use separate prediction heads with lightweight MLPs to predict each target $P\in {\mathbb{R}^{N \times K}}$ based on features produced by previous decoders. 

\begin{figure*}[ht!]
  \centering
  \includegraphics[scale=0.55]{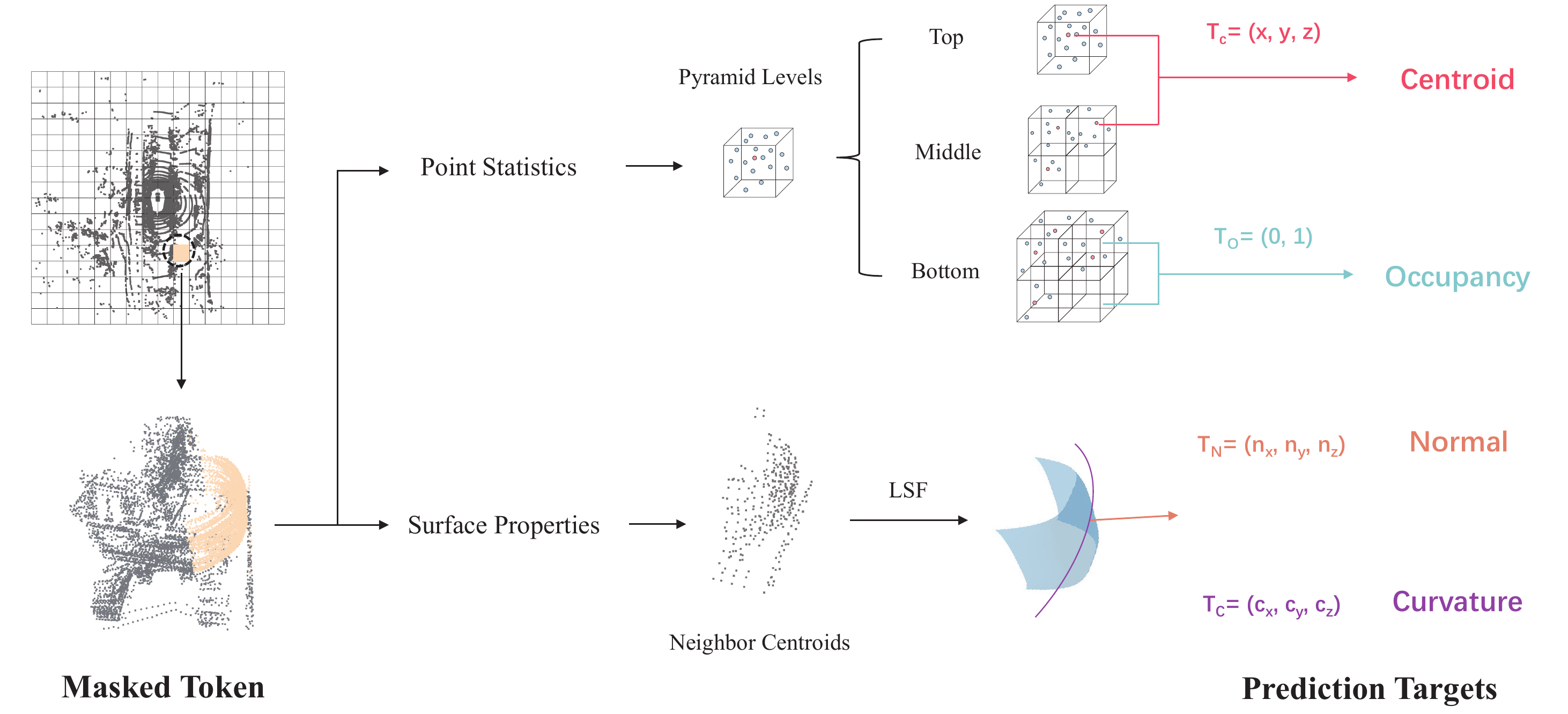}
   \vspace{-5pt}
   \caption{Prediction Targets. We introduce point statistics and surface properties prediction targets to guide the model learning the geometric features of the point cloud. The point statistics targets contain the occupancy of each voxels and centroid of non-empty voxels in a pyramid level. The surface properties prediction targets include surface normal and surface curvature which are obtained by an estimation calculation.}
   \label{fig:prediction targets}
\end{figure*}

\paragraph{Prediction Targets.} 
The prediction targets include the point statistics and surface properties of a point cloud region. The point statistics contain two objectives: pyramid centroid $T_\text{cent}$ and occupancy $T_\text{occ}$. There are also two objectives for surface properties: surface normal $T_\text{norm}$ and surface curvature $T_\text{curv}$. The details of each prediction target will be discussed in Section \ref{sec:voxel-target} and Section \ref{sec:surface-target}. 
\par
We train our network to learn the point statistics and surface properties of uneven point clouds by supervising those prediction targets: 
\begin{equation}\label{eq:loss-separate}
\begin{aligned}
    \mathcal{L}_\text{point} &=\mathcal{L}_\text{cent}(P_\text{cent},T_\text{cent}) + \mathcal{L}_\text{occ}(P_\text{occ},T_\text{occ}), \\
    \mathcal{L}_\text{surface} &=\mathcal{L}_\text{curv}(P_\text{curv},T_\text{curv}) + \mathcal{L}_\text{nor}(P_\text{nor},T_\text{nor}), \\
\end{aligned}
\end{equation}
For centroid, curvature and normal prediction, we use MSE loss, and for occupancy prediction we use Cross-Entropy loss. The overall loss function of our framework is defined as:
\begin{equation}\label{eq:loss-all}
    \mathcal{L}=\mathcal{L}_\text{point}+\mathcal{L}_\text{surface}
\end{equation}

\subsection{Point Statistics Prediction} \label{sec:voxel-target}
Different from 2D images and 3D indoor point clouds, outdoor point clouds are sparse and occluded. Point density varies much in a point cloud, which prevents models directly predicting original point coordinates. The pilot study in Figure~\ref{fig:difference} also shows that such a prediction target is not available. To deal with non-uniform points, we opt to predict  centroid of points in each voxel. In addition, to incorporate multi-scale information, we aim to predict these statistics in different scales by building a voxel pyramid. As shown in Figure~\ref{fig:prediction targets}, we break each masked voxel into three sub-voxel levels (top, middle, and bottom) and compute the voxel occupancy and centroid at each level.

\noindent \textbf{Centroid and Occupancy.}\quad Let $\mathcal{G}^{l}=\{{G}_{i}^{l}=\{ {I}_{{G}_{i}^{l}} \} |i=1,...,{N}_{l}; l\in\{top, middle, bottom \}\}$ be the set of non-empty grids in the l-th pyramid level where ${I}_{{G}_{i}^{l}}$ is the grid index, and ${N}_{l}$ is the number of non-empty grids. Points that are within the same grid ${G}_{i}^{l}$ are grouped together into a set $\mathcal{N}({G}_{i}^{l})$  by calculating their belonging grid index ${I}_{{G}_{i}^{l}}$ from their spatial coordinates. The point centroid of each grid ${G}_{i}^{l}$ is then calculated as:
\begin{equation}\label{eq:centroid}
    c_{G_i^l}=\frac{1}{\lvert \mathcal{N}(G_i^l) \rvert} \sum_{x_{p_j}\in \mathcal{N}(G_i^l)} x_{p_j}
\end{equation}
We also introduce an occupancy prediction target to judge whether a grid is empty or not:
\begin{equation}\label{eq:occupancy}
    o_{G_i^l}=\begin{cases}1,\quad $at least one point in the grid$ \\
    0,\quad otherwise
    \end{cases}
\end{equation}
The point statistics prediction targets for each masked token $v_{j}$ can be formalized as:
\begin{equation}\label{eq:voxel-pre}
    P^j_{cent}=\{c_{G_i^l}\},\quad P^j_{occ}=\{o_{G_i^l}\}
\end{equation}

\begin{table*}[!ht]
    \centering
    \begin{adjustbox}{width=1.99\columnwidth,center}
        \begin{tblr}{Q[m,l,30mm]  Q[m,c,35mm] Q[m,c,35mm] Q[m,c,2mm] Q[m,c,20mm] Q[m,c,20mm]}
            \hline[1pt]
            \SetCell[r=2]{m} Method  &  \SetCell[c=2]{m} Waymo  & & & \SetCell[c=2]{m} nuScenes &   \\
            \cline[0.5pt]{2-3} \cline[0.5pt]{5-6}
                                        &    L1 AP/APH  & L2 AP/APH  & &   mAP  &   NDS       \\
            \hline[1pt]
            Scratch                             & $70.68$/$66.39$   & $64.28$/$60.42$ &  & $50.39$   & $55.04$    \\
            BYOL \cite{byol}                    & $70.15$/$65.72$   & $63.71$/$59.94$ &  & $50.01$   & $54.67$   \\
            PointContrast \cite{pointcontrast}  & $71.73$/$67.28$   & $65.34$/$61.45$ &  & $50.96$   & $55.39$  \\
            SwAV \cite{swav}                    & $71.85$/$67.43$   & $65.41$/$61.63$ &  & $51.57$   & $55.72$   \\
            STRL \cite{STRL}                    & $71.91$/$67.64$   & $65.52$/$61.77$ &  & $51.72$   & $55.84$   \\

            \hline
            \textbf{\Name (Ours)} & \textbf{73.71}/\textbf{70.24}   & \textbf{67.30}/\textbf{63.97} &  &  \textbf{53.77}   & \textbf{57.23}    \\
            \hline[1pt]
        \end{tblr}
    \end{adjustbox}
    \vspace{-5pt}
    \caption{Performances of 3D object detection on the Waymo Open Dataset and nuScenes Dataset validation split.}
    \label{tab:ssl_methods_Waymo_nus}
    \vspace{-5pt}
\end{table*}

\subsection{Surface Properties Prediction} \label{sec:surface-target}
LiDAR point clouds naturally preserve geometric information. Although point statistics provide a rough estimation of shapes, they cannot describe the fine-grained geometric information that are usually critical to recognition tasks. Therefore,  in addition to point statistics prediction, we further leverage 3D shape geometry of point clouds for self-supervised learning. Our desiderata include: these geometric features should be easy to compute and accurately approximate local shape geometry; we can estimate these features from local point groups. Therefore, our choices are surface curvature and surface normals which can be computed in closed-form from local points. To obtain a more stable geometric representation, we incorporate points from 8 neighboring voxels in addition to inside points in each voxel.

\noindent \textbf{Surface Normal and Curvature.}\quad Inspired by surface feature estimation (i.e., curvature estimation\cite{zhang2008curvature} and normal estimation \cite{mitra2003estimating}) in geometry processing, we adopt local least square fitting to handle noisy LiDAR point clouds. Given a set of $K$ gathered points $\mathbf{p}_i$ ($1 \leq i \leq K$), we compute a covariance matrix
\begin{equation}
M=\frac{1}{K} \sum_{i=1}^K \mathbf{p}_i \mathbf{p}_i^T-\bar{\mathbf{p}} \bar{\mathbf{p}}^T,
\end{equation}
where $M$ is a $3\times3$ symmetric matrix, $\bar{\mathbf{p}}$ is the centroid of this point cluster. After the eigen-decomposition of $M$, we obtain eigenvalues $\lambda_1$, $\lambda_2$, $\lambda_3$ ($\lambda_1 \ge \lambda_2 \ge \lambda_3$) and their corresponding eigenvectors and $\mathbf{n}_1$, $\mathbf{n}_2$, $\mathbf{n}_3$. In fact, we use singular value decomposition. Following the aforementioned work~\cite{mitra2003estimating}, the normal vector for each voxel is $\mathbf{n}_3$ (the corresponding eigenvector of the least eigenvalue). Moreover, we compute three pseudo curvature vectors $c_m$  for each point: 
\begin{equation}
c_m = \frac{\lambda_m}{\sum_{i=1}^3 \lambda_i}, m \in \{1, 2, 3\}.
\end{equation}
Therefore, surface properties prediction targets for each masked token $v_{j}$ can be formalized as:
\begin{equation}\label{eq:surface-target}
    P^j_{nor}=n_3^j ,\quad P^j_{curv}=\{c_0^j, c_1^j, c_2^j\}
\end{equation}


\section{Experiments}
In this section, we evaluate our proposed \Name on two widely used benchmarks: Waymo Open Dataset \cite{waymo} and nuScene Dataset \cite{nuscenes}. We first elaborate the experiment setting in Section~\ref{exp_set}. In Section~\ref{exp_previous} we compare our method with previous self-supervised point cloud representation learning methods. In Section~\ref{exp_downstream} we show the generalization of our method on different downstream tasks. In Section~\ref{exp_ablation}, we conduct various ablation studies to evaluate the effectiveness of our approach.

\begin{table*}[ht]
    \centering
    \begin{adjustbox}{width=1.99\columnwidth,center}
        \begin{tblr}{Q[m,l,20mm] Q[m,c,25mm] Q[m,c,30mm] Q[m,c,30mm] Q[m,c,5mm] Q[m,c,18mm] Q[m,c,18mm]}
            \hline[1pt]
            \SetCell[r=2]{m} Method & \SetCell[r=2]{m} Detection Head  &  \SetCell[c=2]{m} Waymo  & & & \SetCell[c=2]{m} nuScenes &   \\
            \cline[0.5pt]{3-4} \cline[0.5pt]{6-7} 
                                &     &    L1 AP/APH  & L2 AP/APH  & &   mAP  &   NDS       \\
            \hline[1pt]
            SpConv$^\ast$  &\SetCell[r=3]{m} Anchor-base    & $66.74$/$63.05$   & $60.27$/$57.12$ & & $48.46$ & $53.92$    \\
            SST    &                                &  $70.68$/$66.39$   & $64.28$/$60.42$ & & $50.39$   & $55.04$   \\
            \textbf{\Name} & & \textbf{73.71}/\textbf{70.24}   & \textbf{67.30}/\textbf{63.97} & & \textbf{53.77}   & \textbf{57.23}  \\
            \hline[1pt]
            SpConv$^\ast$  &\SetCell[r=3]{m} Center-base  & $69.45$/$65.27$    & $63.19$/$59.44$  & & $55.67$   & $64.03$   \\
            SST &                                   & $70.23$/$66.29$    & $64.00$/$60.39$  & & $55.71$   & $64.07$   \\
            \textbf{\Name} &  & \textbf{73.14}/\textbf{69.68}    & \textbf{66.95}/\textbf{63.75}  & & \textbf{58.73}  & \textbf{66.68}   \\
            \hline[1pt]
        \end{tblr}
    \end{adjustbox}
    \caption{Performances of 3D object detection on the Waymo Open Dataset and nuScenes Dataset validation split. $^\ast$:  re-implemented by MMDetection3D.}
    \label{tab:detection_Waymo_nus}
\end{table*}

\begin{table}[ht]
    \centering

    \begin{adjustbox}{width=0.99\columnwidth,center}
        \begin{tblr}{Q[m,l,28mm] Q[m,c,28mm] Q[m,c,28mm] }
            \hline[1pt]
            Method     & mAP & NDS   \\
            \hline[1pt]
            PointPillars~\cite{pointpillars}  &  $30.5$   & $45.3$    \\
            3DSSD~\cite{3dssd}  &  $56.4$   & $42.6$    \\ 
            CBGS~\cite{cbgs} &  $63.3$ & $52.8$        \\ 
            CenterPoint~\cite{centerpoint} & $58.0$   &  $65.5$     \\ 
            VISTA~\cite{vista} & $63.0$   &  $69.8$     \\ 
            Focals Conv~\cite{focalconv}  & $63.8$   &  $70.0$    \\ 
            TransFusion-L~\cite{transfusion}  & $65.5$   &  $70.2$    \\ 
            LargeKernel3D~\cite{largekernel3d}  & $65.4$  &  $70.6$     \\ 
            $\textbf{GeoMAE}^{\dag}$ & \textbf{67.8}   & \textbf{72.5}  \\
            \hline[1pt]
        \end{tblr}
    \end{adjustbox}
    \caption{Performances of 3D object detection on the nuScenes test split. $\dag$ means that we use a multi-stride structure compared to the original single-stride design~\cite{sst}.}
    \label{tab:nuscenes_test}
\end{table}
\vspace{-10pt}

\subsection{Experimental Setup}\label{exp_set}
\noindent \textbf{Waymo Open Dataset.}\quad Waymo Open Dataset \cite{waymo} consists of 798 training sequences and 202 validation sequences. The point cloud scene is collected by a 64-beam LiDAR with around 158k point cloud samples in the training split and 40k point cloud samples in the validation split. For 3D detection, the official evaluation metric includes standard 3D mean Average Precision (mAP) and mAP weighted by heading accuracy (mAPH). These metrics are based on an IoU threshold of 0.7 for vehicles and 0.5 for other categories. These metrics are further broken down into two difficulty levels: L1 for boxes with more than five LiDAR points and L2 for boxes with at least one LiDAR point.

\noindent \textbf{nuScenes Dataset.}\quad The nuScenes Dataset \cite{nuscenes} is a large-scale autonomous driving dataset that contains 700, 150, and 150 sequences for training, validation, and testing, respectively. For 3D detection, the major official metrics are mean Average Precision (mAP) and nuScenes detection score (NDS). The mAP uses a bird-eye-view center distance threshold (0.5m, 1m, 2m, 4m) instead of bounding box IoU. NDS is a weighted average of mAP and other attribute metrics, including translation, scale, orientation, velocity, and other box attributes. For 3D tracking, nuScenes mainly uses AMOTA, which penalizes ID switches, false positives, and false negatives and is averaged among various recall thresholds.

\noindent \textbf{Model.}\quad Our proposed \Name use a standard SST \cite{sst} as the encoder, which contains two consecutive blocks. Each attention module in the encoder has two heads, 128 input channels, and 256 hidden channels. We use two parallel decoders, and each decoder has two transformer blocks. Given the masked voxels ${V_{m}}$ with the voxel grid size ${g_{x}}$, ${g_{y}}$, ${g_{z}}$ along the x, y, and z axes of the 3D space, respectively, we sub-divide each voxel's spatial space into three pyramid levels (as shown in Figure~\ref{fig:prediction targets}): top, middle, and bottom. The grid in the top level has the same grid size as the original voxel grid. The sub-grids in the middle level have a grid size of ${g_{x}/2}$, ${g_{y}/2}$, and ${g_{z}/4}$, while the grid size in the bottom level is ${g_{x}/4}$, ${g_{y}/4}$, and ${g_{z}/8}$. The number of grids in each level, from top to bottom, is 1, 16, and 128, respectively.

\noindent \textbf{Training Details.}\quad On the Waymo Open Dataset, we use all the samples for pre-training and uniformly sample 20\% of frames for finetuning following common practice. On the nuScenes Dataset, we use all the frames (including the unlabeled sweeps) for pre-training and the labeled samples for finetuning. 
For pre-training, all the self-supervised methods are trained for 72 epochs (denoted as 6x) with the AdamW optimizer~\cite{adam}. The initial learning rate is 1e-5.
For finetuning downstream tasks, we follow the original training settings in each downstream approach to finetune the detector, segmenter, and tracker.

\subsection{Comparison on 3D Object Detection}\label{exp_previous}
Unlike 2D MIM methods which adopt image classification task as the benchmark to evaluate the effectiveness of their pre-training methods, we do not have a classification task for scene-level 3D point clouds. So we choose the 3D object detection task to compare our method with previous 3D self-supervised methods.

\noindent \textbf{Settings.}\quad
We compare our \Name with several typical 3D self-supervised learning methods including PointContrast \cite{pointcontrast}, STRL \cite{STRL}, BYOL \cite{byol}, and SwAV \cite{swav}. We follow the strategy in \cite{once} to apply these methods for pre-training the SST backbone. Details are presented in the Appendix.
After pre-training, we evaluate the pre-trained backbones based on the 3D object detector benchmark proposed in SECOND \cite{second}. Detectors with different pre-trained SST backbones are all finetuned for 24 epochs on the Waymo Open Dataset and 20 epochs on the nuScenes Dataset.

\noindent \textbf{Results.}\quad
The finetuning results on 3D object detection are shown in Table~\ref{tab:ssl_methods_Waymo_nus}. Our proposed \Name  significantly improves SST, which is 3.03/3.85 L1 AP/APH better than training from scratch on the Waymo Open Dataset and 3.38 mAP on the nuScenes Dataset.
Compared with other self-supervised methods, \Name outperforms the second best method STRL~\cite{STRL} by a significant margin, 1.78/2.20 L1 AP/APH on the Waymo Open Dataset and 2.05 mAP on the nuScenes Dataset, demonstrating the effectiveness of our model and prediction target designs.

\subsection{Comparison on other Downstream Tasks}\label{exp_downstream}
We further evaluate the effectiveness and generalization of \Name in different 3D downstream tasks (including detection, segmentation and tracking) with different model architectures (backbones and heads). 
For each model in a downstream task, we evaluate three variants with different backbones: 1. the original sparse convolutional networks without pre-training; 2. SST without pre-training; 3. SST pre-trained by our \Name. 

\begin{table*}[!ht]
    \centering
    \begin{adjustbox}{width=1.99\columnwidth,center}
        \begin{tblr}{Q[m,l,35mm] Q[m,c]  Q[m,c] Q[m,c] Q[m,c] Q[m,c] Q[m,c] Q[m,c] Q[m,c] Q[m,c] Q[m,c] Q[m,c] Q[m,c] Q[m,c] Q[m,c] Q[m,c] Q[m,c] Q[m,c]  }
            \hline[1pt]
        Methods &  mIoU & \rotatebox{90}{barrier} & \rotatebox{90}{bicycle} & \rotatebox{90}{bus} & \rotatebox{90}{car} & \rotatebox{90}{construction} & \rotatebox{90}{motorcycle} & \rotatebox{90}{pedestrian} & \rotatebox{90}{traffic-cone} & \rotatebox{90}{trailer} & \rotatebox{90}{truck} & \rotatebox{90}{driveable} & \rotatebox{90}{other}
        & \rotatebox{90}{sidewalk} & \rotatebox{90}{terrain} & \rotatebox{90}{manmade} & \rotatebox{90}{vegetation} \\
            \hline[1pt]
            Cylinder3D \cite{cylinder3d}    & $76.1$ & $76.4$ & $40.3$ & $91.2$ & $93.8$ & $51.3$ & $78.0$ & $78.9$ & $64.9$ & $62.1$ & $84.4$ & $96.8$ & $71.6$ & $76.4$ & $75.4$ & $90.5$ & $87.4$     \\
            Cylinder3D-SST    & $76.5$ & $76.2$ & $40.0$ & $91.8$ & $94.2$ & $51.6$ & $78.1$ & $80.1$ & $64.7$ & $62.5$ & $84.7$ & $97.1$ & $71.7$ & $76.7$ & $75.8$ & $90.8$ & $87.7$   \\
            \textbf{Cylinder3D-\Name}    & \textbf{78.6} & \textbf{78.2} & \textbf{42.6} & \textbf{93.5} & \textbf{95.8} & \textbf{55.4} & \textbf{79.8} & \textbf{83.5} & \textbf{66.8} & \textbf{65.6} & \textbf{87.3} & \textbf{97.7} & \textbf{73.3} & \textbf{78.2} & \textbf{77.4} & \textbf{92.6} & \textbf{89.5}  \\
            \hline[1pt]
        \end{tblr}
    \end{adjustbox}
    \vspace{-5pt}
    \caption{Performances of 3D semantic segmentation on the nuScenes Dataset validation split.}
    \label{tab:segmentation_nus}
\end{table*}

\begin{table*}[!ht]
    \centering
    \begin{adjustbox}{width=1.99\columnwidth,center}
        \begin{tblr}{Q[m,l,35mm] Q[m,c,10mm] Q[m,c,20mm] Q[m,c,20mm] Q[m,c,20mm] Q[m,c,20mm] Q[m,c,20mm] Q[m,c,20mm] Q[m,c,20mm] }
            \hline[1pt]
        Methods & Modality &  mIoU  & Drivable & Ped. Cross. & Walkway & Stop Line & Carpark & Divider  \\
            \hline[1pt]
            CenterPoint \cite{centerpoint} & \SetCell[r=3]{m} L & $ 48.6$ & $75.6$ & $48.4$ & $57.5$ & $36.5$ & $31.7$ & $41.9$    \\
            CenterPoint-SST   & & $49.7$ & $77.2$ & $49.5$ & $58.7$ & $37.2$ & $32.5$ & $43.1$  \\
            \textbf{CenterPoint-\Name} & & \textbf{52.4}  & \textbf{79.5} & \textbf{53.1} & \textbf{61.6} & \textbf{39.7} & \textbf{34.9} & \textbf{45.6}     \\
            \hline[1pt]
            BEVFusion \cite{bevfusion} & \SetCell[r=3]{m} C + L & $62.7$ & $85.5$ & $60.5$ & $ 67.6$ & $52.0$ & $57.0$  & $53.7$   \\
            BEVFusion-SST       &   & $63.1$ & $86.1$ & $60.8$ & $68.7$  & $53.5$ & $53.8$  & $55.5$    \\
            \textbf{BEVFusion-\Name}       &  & \textbf{65.2} & \textbf{86.5} & \textbf{62.3} & \textbf{70.2} & \textbf{55.7} & \textbf{59.4} & \textbf{56.8}  \\
            \hline[1pt]
        \end{tblr}
    \end{adjustbox}
    \vspace{-5pt}
    \caption{Performances of BEV map segmentation on the nuScenes Dataset validation split.}
    \label{tab:bev_seg_nus}
    \vspace{-6pt}
\end{table*}

\subsubsection{3D Object Detection}
\noindent \textbf{Settings.}\quad
We comprehensively evaluate \Name on both the anchor-based detector SECOND and a center-based detector CenterPoint \cite{centerpoint}. For the Waymo Open Dataset, the detection point cloud range is set to [-74.88m, 74.88m] for X- and Y-axes, [-2m, 4m] for Z-axes, and the voxel size is set to (0.32m, 0.32m, 6m). For nuScenes Dataset, the detection range is set to [-51.2m, 51.2m] for X- and Y-axes, [-5m, 3m] for Z-axes, and the voxel size is set to (0.256m, 0.256m, 8m).

\noindent \textbf{Results.}\quad
As shown in Table~\ref{tab:detection_Waymo_nus}, both anchor-based and center-based detectors with pre-trained SST by our \Name achieve better performance than the baselines. For anchor-based detector, our \Name outperforms the baseline by 3.03 L1 AP on the Waymo Open Dataset and 3.38 mAP on the nuScenes Dataset. While for the center-based detector, our approach improves the results of training from scratch by 2.91 L1 AP on the Waymo Open Dataset and 3.02 mAP on the nuScenes Dataset. All the results verify the efficacy of our proposed method. In Table~\ref{tab:nuscenes_test}, we also test our method on the nuScenes test split and achieve a new state-of-the-art result.
\vspace{3pt}
\vspace{-12pt}
\subsubsection{3D Object Tracking}
\noindent \textbf{Settings.}\quad
We also conduct experiments in a 3D multi-object tracking (MOT) task on the nuScenes Dataset by performing tracking-by-detection algorithms proposed by CenterPoint \cite{centerpoint} and SimpleTrack \cite{simpletrack}. The point cloud range and voxel size are the same as the 3D object detection settings.

\noindent \textbf{Results.}\quad
From Table~\ref{tab:tracking_nus}, we can see that our GeoMAE outperforms the baseline (SST) by 1.7 AMOTA for Centerpoint and 1.1 AMOTA for SimpleTrack.
These observations are consistent with those in 3D object detection.

\begin{table}[ht]
    \centering
    \begin{adjustbox}{width=0.99\columnwidth,center}
        \begin{tblr}{Q[m,l,25mm] Q[m,c,16mm] Q[m,c,16mm] Q[m,c,16mm] Q[m,c,16mm] }
            \hline[1pt]
                   Method   &    AMOTA↑ &   AMOTP↓   & MOTA↑  & IDS↓   \\
            \hline[1pt]
            Centerpoint$^\ast$~\cite{centerpoint}       & $57.3$   & $0.681$ & $0.522$ & $594$    \\
            Centerpoint-SST                             & $59.9$   & $0.660$ & $0.514$ & $586$    \\
            \textbf{Centerpoint-\Name}                & \textbf{61.6}   & \textbf{0.635} & \textbf{0.635} & \textbf{582} \\
            \hline[1pt]
            SimpleTrack$^\ast$~\cite{simpletrack}   & $63.2$   & $0.678$  & $0.548$ & $520$ \\
            SimpleTrack-SST                         & $63.8$   & $0.653$ & $0.541$ & $514$   \\
            \textbf{SimpleTrack-\Name}           & \textbf{64.9}   & \textbf{0.624} & \textbf{0.561} & \textbf{473}    \\
            \hline[1pt]
        \end{tblr}
    \end{adjustbox}
    \vspace{-5pt}
    \caption{Performances of 3D multi-object tracking on the nuScenes Dataset validation split. $^\ast$: re-implemented by MMDetection3D.}
    \label{tab:tracking_nus}
    \vspace{-6pt}
\end{table}

\subsubsection{LiDAR Semantic Segmentation}
\noindent \textbf{Settings.}\quad
To demonstrate the generalization capability, we evaluate our method on the nuScenes Dataset for the LiDAR segmentation task. We follow the official guidance to leverage mean intersection-over-union (mIoU) as the evaluation metric. We adopt the Cylinder3D \cite{cylinder3d} as our baseline architecture and replace the last stage of the backbone from sparse convolutions into SST. Other training settings are the same as in Cylinder3D.
\par
\noindent \textbf{Results.}\quad
As reported in Table~\ref{tab:segmentation_nus}, the Cylinder3D obtains $0.4$ performance gain by replacing the backbone from sparse convolutions with our modified SST.
When applying our \Name to pre-train the backbone, it achieves $2.1$ mIoU gain than training from scratch. 

\subsubsection{BEV Map Segmentation}
\noindent \textbf{Settings.}\quad
We further experiment our method in the BEV Map Segmentation task on the nuScenes Dataset. We perform the evaluation in the [-50m, 50m]×[-50m, 50m] region following the common practice in BEVFusion~\cite{bevfusion}. We develop the CenterPoint-SST and BEVFusion-SST by replacing the last two stages of the LiDAR backbone with SST.
\par
\noindent \textbf{Results.}\quad
We report the BEV map segmentation results in Table~\ref{tab:bev_seg_nus}. For the LiDAR-only model, our method surpasses the SST baseline by 2.7 mIoU. In the multi-modality setting, \Name further boosts the performance of BEVFusion-SST about 2.1 mIoU, which demonstrates the strong generalization capability of our method.

\subsection{Ablation Study}\label{exp_ablation}

We adopt standard SST \cite{sst} as the default backbone in our ablation study. To get efficient validation and reduce experimental overhead, all the experiments are pre-trained for 72 (6x) epochs if not specified.

\begin{table}[!ht]
    \centering
    \begin{adjustbox}{width=0.99\columnwidth,center}
        \begin{tblr}{Q[m,l,30mm] Q[m,c,20mm] Q[m,c,20mm] Q[m,c,20mm]}
            \hline[1pt]
            Methods      &  Type & Waymo L1 AP & nuScenes mAP    \\
            \hline[1pt]
            Scratch          &                                          & $70.68$  & $50.39$    \\
            \hline[1pt]
            + Centroid  & \SetCell[r=2]{m}  Point Statistics            & $71.60$   & $51.25$   \\
            + Occupancy                  &                              & $72.65$   & $52.12$  \\
            \hline[1pt]
            + Surface Normal    & \SetCell[r=2]{m} Surface Properties    & $73.37$   & $52.94$    \\
            + Surface Curvature          &                               & \textbf{73.71}   & \textbf{53.31}   \\
            \hline[1pt]
        \end{tblr}
    \end{adjustbox}
    \vspace{-5pt}
    \caption{Ablation study on prediction targets. Detection results on the Waymo and nuScenes Dataset.}
    \label{tab:ablation_componet}
    \vspace{-5pt}
\end{table}

\noindent \textbf{Prediction Targets.}\quad
We present ablation studies in Table~\ref{tab:ablation_componet} to justify our design choices. Scratch means training on the detection from scratch without pre-training the backbone. From the table we can see that by adopting centroid and occupancy as the prediction targets, the model performs better than training form scratch with about 1.87 L1 AP gain on the Waymo Open Dataset and 1.7 mAP gain on the nuScene Dataset. 
As shown in the last two rows, we investigate the effect of predicting the surface normal and curvature. The additional prediction targets further improve the performance 1.06 AP on the Waymo Open Dataset and 1.19 mAP on the nuScenes Dataset. 

\noindent \textbf{Decoder Design.}\quad
The separate decoder is one key component in our \Name. We ablate the design in Table~\ref{tab:ablation_decoder}. Shared means we use a single decoder to decode both Point Statistics and Surface Properties information. Same target means we use two decoders but the separate decoders decode and predict the same Point Statistics targets or the same Surface Properties targets. It can be observed that our final design achieves the best performance, which indicates that such separate decoder truly disentangles the different representations and enable the model to learn different geometry information.
\begin{table}[!h]
    \centering
    \begin{adjustbox}{width=0.99\columnwidth,center}
        \begin{tblr}{l c c  }
            \hline[1pt]
            Decoder       & Waymo L1 AP & nuScenes mAP   \\
            \hline[1pt]
            Shared Decoder                              & $72.45$   & $52.04$  \\
            Separate (Only Point Statistics)            & $72.65$   & $52.28$   \\
            Separate (Only Surface Properties)          & $72.03$   & $51.87$  \\
            Separate (Different Targets)                & \textbf{73.71}   & \textbf{53.31}  \\
            \hline[1pt]
        \end{tblr}
    \end{adjustbox}
    \vspace{-5pt}
    \caption{Ablation study on decoder design. Detection results on the Waymo and nuScenes Dataset.}
    \label{tab:ablation_decoder}
\end{table}

\noindent \textbf{Masking Ratio.}\quad
We study the influence of the masking ratio in Table~\ref{tab:ablation_masking_ratio}. Similar to the observation on images, the optimal masking ratio is high (70\%). The performance degrades largely with too low or too high masking ratios.

\begin{table}[ht]
    \centering
    \begin{adjustbox}{width=0.99\columnwidth,center}
        \begin{tblr}{Q[m,c,30mm] Q[m,c,30mm] Q[m,c,30mm]}
            \hline[1pt]
            Training Epochs   & Waymo L1 AP & nuScenes mAP   \\
            \hline[1pt]
            24        & $72.67$   & $52.64$   \\
            48        & $73.32$   & $52.97$   \\
            72        & \textbf{73.71}   & \textbf{53.31}  \\
            96        & $73.71$   & $53.30$  \\
            \hline[1pt]
        \end{tblr}
    \end{adjustbox}
    \vspace{-5pt}
    \caption{Ablation on training schedule. Detection results on the Waymo and nuScenes Dataset.}
    \label{tab:ablation_schedule}
\end{table}

\begin{table}[ht]
    \centering
    \begin{adjustbox}{width=0.99\columnwidth,center}
        \begin{tblr}{Q[m,c,30mm] Q[m,c,30mm] Q[m,c,30mm]}
            \hline[1pt]
            Masking Ratio       & Waymo L1 AP & nuScenes mAP   \\
            \hline[1pt]
            40\%        & $72.27$   & $52.47$   \\
            60\%        & $73.28$   & $53.00$   \\
            70\%        & \textbf{73.71}   & \textbf{53.31}  \\
            80\%        & $71.94$   & $51.85$  \\
            \hline[1pt]
        \end{tblr}
    \end{adjustbox}
    \vspace{-5pt}
    \caption{Ablation on masking ratio. Detection results on the Waymo and nuScenes Dataset.}
    \label{tab:ablation_masking_ratio}
\end{table}

\begin{table}[ht]
    \centering
    \begin{adjustbox}{width=0.99\columnwidth,center}
        \begin{tblr}{Q[m,c,30mm] Q[m,c,30mm] Q[m,c,30mm]}
            \hline[1pt]
            Dataset Scale      & Waymo L1 AP & nuScenes mAP   \\
            \hline[1pt]
            0\%         & $70.68$   & $50.39$ \\
            20\%        & $72.27$   & $51.64$   \\
            50\%        & $72.88$   & $52.74$   \\
            80\%        & $73.48$   & $53.03$  \\
            100\%       & \textbf{73.71}   & \textbf{53.31} \\
            \hline[1pt]
        \end{tblr}
    \end{adjustbox}
    \vspace{-5pt}
    \caption{Impacts of different scale of the pre-training dataset. Detection results on the Waymo and nuScenes Dataset.}
    \label{tab:ablation_dataset_scale}
\end{table}

\noindent \textbf{Training Schedule.}\quad 
Table~\ref{tab:ablation_schedule} shows the effect of the training schedule length. We ablate the pre-training schedule of \Name from 24 (2x) to 96 (8x) epochs and fix the fine-tuning epoch as 24 epochs. The accuracy improves steadily with longer training schedule until 72 epochs.

\noindent \textbf{Pre-training Dataset Scale.}\quad
We also investigate the effect of different scales of  pre-training dataset. As shown in Table~\ref{tab:ablation_dataset_scale}, performance grows as the scale of pre-training data increases. And our method still achieves about 1 point performance improvements on both datasets, which indicates the effectiveness of our GeoMAE.  
\vspace{-5pt}

\section{Conclusion}

We present GeoMAE, a geometry-aware self-supervised pre-training approach for point clouds. GeoMAE achieves strong performance on a variety of downstream tasks including 3D detection, segmentation, and tracking. GeoMAE leverages recent development in masked modeling. In addition to the commonly used occupancy prediction target, our method proposes three additional learning objectives, which jointly become a challenging and informative pretext task. Our key observation is that geometric features provide strong information for models to reason objects and scenes, therefore improving downstream recognition performance. Our results also suggest several venues for future inquiry. First, pre-training using GeoMAE on a larger unlabeled dataset will further boost the performance (\eg, pre-training on DDAD dataset~\cite{packnet}). Besides, exploring other types of geometric features remains an open and intriguing question. 

\paragraph{Acknowledgement} This work was supported by National Key R\&D Program of China (2022ZD0161700).

\section{Appendix}
We reproduce four previous self-supervised learning methods, including two contrastive learning methods tailored to point clouds (PointContrast~\cite{pointcontrast} and STRL~\cite{STRL}), as well as two typical self-supervised learning methods (BYOL~\cite{byol} and SwAV~\cite{swav} ). 

\noindent \textbf{General Configurations.}\quad We adopt the standard SST as the backbone. For the Waymo Open Dataset~\cite{waymo}, the point cloud range is set to [-74.88m, 74.88m] for X-axes and Y-axes, [-2m, 4m] for Z-axes, and the voxel size is set to (0.32m, 0.32m, 6m). For nuScenes Dataset~\cite{nuscenes}, the point cloud range is set to [-51.2m, 51.2m] for X-axes and Y-axes, [-5m, 3m] for Z-axes, and the voxel size is set to (0.256m, 0.256m, 8m). For all the methods, the pretaining learning rate is initialized as 1e-5, and the fine-tuning learning rate is  
initialized as 1e-4. We use the Adam optimizer and the cosine annealing learning scheme. The models are trained
with batch size 64.

\noindent \textbf{PointContrast.}\quad
We first transform the original point cloud into two augmented views by random geometric transformations, which include random flip, random scaling with a scale factor sampled uniformly from [0.95, 1.05] and random rotation around vertical yaw axis by an angle between [-15, 15] degrees. The scenes will be passed through the SST backbone to obtain voxel-wise features. We randomly select half of the voxel features and then embed them into latent space by using a two-layer MLP (with BatchNorm and ReLU, and the dimensions are 128, 64). The latent space feature will be concatenated with initial features and passed through a one-layer MLP with dimension 64. The concatenated features are used for comparative learning as in the original PointContrast.

\noindent \textbf{BYOL.}\quad
BYOL consists of two networks, an online network and a target network. It iteratively bootstraps the outputs of the target network to serve as targets without using negative pairs. We train its online network to predict the target network’s representation of the other augmented view of the same 3D scene. We pass the voxel-wise features through a two-layer MLP (with dimensions 512, 2048). After that, a two-layer MLP (with dimensions 4096, 256) predictor in the online network will project the embeddings into a latent space as the final representation of the online network. The target network is updated by a slow-moving averaging of the online network with the parameter 0.999. For other configurations, we follow the settings in the original paper.

\noindent \textbf{SwAV.}\quad
Different from contrastive learning methods, SwAV does not directly compare embedding features by introducing prototypes and swapped predictions. Similar to the implementation of PointContrast, we apply the same view generation module and obtain voxel-vise features of different views. We adopt a two-layer MLP projection head with dimensions 512 and 128. We then compute “codes” by assigning features to prototype vectors. Note that we do not adopt the multi-crop strategy proposed in the original paper due to the differences between images and point clouds.

\noindent \textbf{STRL.}\quad STRL learns invariant
representations from two augmented views, which are obtained by spatial augmentation and temporal sampling. For spatial data augmentation, we adopt the same generation approach in PointContrast. For temporal sampling, we follow the settings in the original paper. We add a max-pooling layer at the end of the backbone to obtain the global features. The global features are passed through a projector and a predictor for contrastive learning.

{\small
\bibliographystyle{ieee_fullname}
\bibliography{egbib}

\begin{thebibliography}{10}\itemsep=-1pt

\bibitem{data2vec}
Alexei Baevski, Wei-Ning Hsu, Qiantong Xu, Arun Babu, Jiatao Gu, and Michael
  Auli.
\newblock Data2vec: A general framework for self-supervised learning in speech,
  vision and language.
\newblock {\em arXiv preprint arXiv:2202.03555}, 2022.

\bibitem{transfusion}
Xuyang Bai, Zeyu Hu, Xinge Zhu, Qingqiu Huang, Yilun Chen, Hongbo Fu, and
  Chiew-Lan Tai.
\newblock Transfusion: Robust lidar-camera fusion for 3d object detection with
  transformers.
\newblock In {\em Proceedings of the IEEE/CVF Conference on Computer Vision and
  Pattern Recognition}, pages 1090--1099, 2022.

\bibitem{beit}
Hangbo Bao, Li Dong, and Furu Wei.
\newblock Beit: Bert pre-training of image transformers.
\newblock {\em arXiv preprint arXiv:2106.08254}, 2021.

\bibitem{nuscenes}
Holger Caesar, Varun Bankiti, Alex~H Lang, Sourabh Vora, Venice~Erin Liong,
  Qiang Xu, Anush Krishnan, Yu Pan, Giancarlo Baldan, and Oscar Beijbom.
\newblock nuscenes: A multimodal dataset for autonomous driving.
\newblock In {\em Proceedings of the IEEE/CVF conference on computer vision and
  pattern recognition}, pages 11621--11631, 2020.

\bibitem{swav}
Mathilde Caron, Ishan Misra, Julien Mairal, Priya Goyal, Piotr Bojanowski, and
  Armand Joulin.
\newblock Unsupervised learning of visual features by contrasting cluster
  assignments.
\newblock {\em Advances in Neural Information Processing Systems},
  33:9912--9924, 2020.

\bibitem{cae}
Xiaokang Chen, Mingyu Ding, Xiaodi Wang, Ying Xin, Shentong Mo, Yunhao Wang,
  Shumin Han, Ping Luo, Gang Zeng, and Jingdong Wang.
\newblock Context autoencoder for self-supervised representation learning.
\newblock {\em arXiv preprint arXiv:2202.03026}, 2022.

\bibitem{focalconv}
Yukang Chen, Yanwei Li, Xiangyu Zhang, Jian Sun, and Jiaya Jia.
\newblock Focal sparse convolutional networks for 3d object detection.
\newblock In {\em Proceedings of the IEEE/CVF Conference on Computer Vision and
  Pattern Recognition}, pages 5428--5437, 2022.

\bibitem{largekernel3d}
Yukang Chen, Jianhui Liu, Xiaojuan Qi, Xiangyu Zhang, Jian Sun, and Jiaya Jia.
\newblock Scaling up kernels in 3d cnns.
\newblock {\em arXiv preprint arXiv:2206.10555}, 2022.

\bibitem{Minkowski}
Christopher Choy, JunYoung Gwak, and Silvio Savarese.
\newblock 4d spatio-temporal convnets: Minkowski convolutional neural networks.
\newblock In {\em Proceedings of the IEEE/CVF Conference on Computer Vision and
  Pattern Recognition}, pages 3075--3084, 2019.

\bibitem{hog}
Navneet Dalal and Bill Triggs.
\newblock Histograms of oriented gradients for human detection.
\newblock In {\em 2005 IEEE computer society conference on computer vision and
  pattern recognition (CVPR'05)}, volume~1, pages 886--893. Ieee, 2005.

\bibitem{vista}
Shengheng Deng, Zhihao Liang, Lin Sun, and Kui Jia.
\newblock Vista: Boosting 3d object detection via dual cross-view spatial
  attention.
\newblock In {\em Proceedings of the IEEE/CVF Conference on Computer Vision and
  Pattern Recognition}, pages 8448--8457, 2022.

\bibitem{bert}
Jacob Devlin, Ming-Wei Chang, Kenton Lee, and Kristina Toutanova.
\newblock Bert: Pre-training of deep bidirectional transformers for language
  understanding.
\newblock {\em arXiv preprint arXiv:1810.04805}, 2018.

\bibitem{peco}
Xiaoyi Dong, Jianmin Bao, Ting Zhang, Dongdong Chen, Weiming Zhang, Lu Yuan,
  Dong Chen, Fang Wen, and Nenghai Yu.
\newblock Peco: Perceptual codebook for bert pre-training of vision
  transformers.
\newblock {\em arXiv preprint arXiv:2111.12710}, 2021.

\bibitem{3dssl-gan}
Benjamin Eckart, Wentao Yuan, Chao Liu, and Jan Kautz.
\newblock Self-supervised learning on 3d point clouds by learning discrete
  generative models.
\newblock In {\em Proceedings of the IEEE/CVF Conference on Computer Vision and
  Pattern Recognition}, pages 8248--8257, 2021.

\bibitem{sst}
Lue Fan, Ziqi Pang, Tianyuan Zhang, Yu-Xiong Wang, Hang Zhao, Feng Wang, Naiyan
  Wang, and Zhaoxiang Zhang.
\newblock {Embracing Single Stride 3D Object Detector with Sparse Transformer}.
\newblock In {\em CVPR}, 2022.

\bibitem{byol}
Jean-Bastien Grill, Florian Strub, Florent Altch{\'e}, Corentin Tallec, Pierre
  Richemond, Elena Buchatskaya, Carl Doersch, Bernardo Avila~Pires, Zhaohan
  Guo, Mohammad Gheshlaghi~Azar, et~al.
\newblock Bootstrap your own latent-a new approach to self-supervised learning.
\newblock {\em Advances in neural information processing systems},
  33:21271--21284, 2020.

\bibitem{packnet}
Vitor Guizilini, Rares Ambrus, Sudeep Pillai, Allan Raventos, and Adrien
  Gaidon.
\newblock 3d packing for self-supervised monocular depth estimation.
\newblock In {\em IEEE Conference on Computer Vision and Pattern Recognition
  (CVPR)}, 2020.

\bibitem{mae}
Kaiming He, Xinlei Chen, Saining Xie, Yanghao Li, Piotr Doll{\'a}r, and Ross
  Girshick.
\newblock Masked autoencoders are scalable vision learners.
\newblock In {\em Proceedings of the IEEE/CVF Conference on Computer Vision and
  Pattern Recognition}, pages 16000--16009, 2022.

\bibitem{hoppe1992surface}
Hugues Hoppe, Tony DeRose, Tom Duchamp, John McDonald, and Werner Stuetzle.
\newblock Surface reconstruction from unorganized points.
\newblock In {\em Proceedings of the 19th annual conference on computer
  graphics and interactive techniques}, pages 71--78, 1992.

\bibitem{STRL}
Siyuan Huang, Yichen Xie, Song-Chun Zhu, and Yixin Zhu.
\newblock Spatio-temporal self-supervised representation learning for 3d point
  clouds.
\newblock In {\em Proceedings of the IEEE/CVF International Conference on
  Computer Vision}, pages 6535--6545, 2021.

\bibitem{adam}
Diederik~P Kingma and Jimmy Ba.
\newblock Adam: A method for stochastic optimization.
\newblock {\em arXiv preprint arXiv:1412.6980}, 2014.

\bibitem{pointpillars}
Alex~H Lang, Sourabh Vora, Holger Caesar, Lubing Zhou, Jiong Yang, and Oscar
  Beijbom.
\newblock Pointpillars: Fast encoders for object detection from point clouds.
\newblock In {\em Proceedings of the IEEE/CVF conference on computer vision and
  pattern recognition}, pages 12697--12705, 2019.

\bibitem{sonet}
Jiaxin Li, Ben~M Chen, and Gim~Hee Lee.
\newblock So-net: Self-organizing network for point cloud analysis.
\newblock In {\em Proceedings of the IEEE conference on computer vision and
  pattern recognition}, pages 9397--9406, 2018.

\bibitem{a2mim}
Siyuan Li, Di Wu, Fang Wu, Zelin Zang, Baigui Sun, Hao Li, Xuansong Xie, Stan
  Li, et~al.
\newblock Architecture-agnostic masked image modeling--from vit back to cnn.
\newblock {\em arXiv preprint arXiv:2205.13943}, 2022.

\bibitem{pointcnn}
Yangyan Li, Rui Bu, Mingchao Sun, Wei Wu, Xinhan Di, and Baoquan Chen.
\newblock Pointcnn: Convolution on x-transformed points.
\newblock {\em Advances in neural information processing systems}, 31, 2018.

\bibitem{point2sequence}
Xinhai Liu, Zhizhong Han, Yu-Shen Liu, and Matthias Zwicker.
\newblock Point2sequence: Learning the shape representation of 3d point clouds
  with an attention-based sequence to sequence network.
\newblock In {\em Proceedings of the AAAI Conference on Artificial
  Intelligence}, volume~33, pages 8778--8785, 2019.

\bibitem{swin-transformer}
Ze Liu, Yutong Lin, Yue Cao, Han Hu, Yixuan Wei, Zheng Zhang, Stephen Lin, and
  Baining Guo.
\newblock Swin transformer: Hierarchical vision transformer using shifted
  windows.
\newblock In {\em Proceedings of the IEEE/CVF International Conference on
  Computer Vision}, pages 10012--10022, 2021.

\bibitem{bevfusion}
Zhijian Liu, Haotian Tang, Alexander Amini, Xinyu Yang, Huizi Mao, Daniela Rus,
  and Song Han.
\newblock Bevfusion: Multi-task multi-sensor fusion with unified bird's-eye
  view representation.
\newblock {\em arXiv preprint arXiv:2205.13542}, 2022.

\bibitem{once}
Jiageng Mao, Minzhe Niu, Chenhan Jiang, Hanxue Liang, Jingheng Chen, Xiaodan
  Liang, Yamin Li, Chaoqiang Ye, Wei Zhang, Zhenguo Li, et~al.
\newblock One million scenes for autonomous driving: Once dataset.
\newblock {\em arXiv preprint arXiv:2106.11037}, 2021.

\bibitem{voxnet}
Daniel Maturana and Sebastian Scherer.
\newblock Voxnet: A 3d convolutional neural network for real-time object
  recognition.
\newblock In {\em 2015 IEEE/RSJ international conference on intelligent robots
  and systems (IROS)}, pages 922--928. IEEE, 2015.

\bibitem{mitra2003estimating}
Niloy~J Mitra and An Nguyen.
\newblock Estimating surface normals in noisy point cloud data.
\newblock In {\em Proceedings of the nineteenth annual symposium on
  Computational geometry}, pages 322--328, 2003.

\bibitem{point-mae}
Yatian Pang, Wenxiao Wang, Francis~EH Tay, Wei Liu, Yonghong Tian, and Li Yuan.
\newblock Masked autoencoders for point cloud self-supervised learning.
\newblock {\em arXiv preprint arXiv:2203.06604}, 2022.

\bibitem{simpletrack}
Ziqi Pang, Zhichao Li, and Naiyan Wang.
\newblock Simpletrack: Understanding and rethinking 3d multi-object tracking.
\newblock {\em arXiv preprint arXiv:2111.09621}, 2021.

\bibitem{pointnet}
Charles~R Qi, Hao Su, Kaichun Mo, and Leonidas~J Guibas.
\newblock Pointnet: Deep learning on point sets for 3d classification and
  segmentation.
\newblock In {\em Proceedings of the IEEE conference on computer vision and
  pattern recognition}, pages 652--660, 2017.

\bibitem{pointnet++}
Charles~Ruizhongtai Qi, Li Yi, Hao Su, and Leonidas~J Guibas.
\newblock Pointnet++: Deep hierarchical feature learning on point sets in a
  metric space.
\newblock {\em Advances in neural information processing systems}, 30, 2017.

\bibitem{repsurf}
Haoxi Ran, Jun Liu, and Chengjie Wang.
\newblock Surface representation for point clouds.
\newblock In {\em Proceedings of the IEEE/CVF Conference on Computer Vision and
  Pattern Recognition}, pages 18942--18952, 2022.

\bibitem{3dssl-global-local}
Yongming Rao, Jiwen Lu, and Jie Zhou.
\newblock Global-local bidirectional reasoning for unsupervised representation
  learning of 3d point clouds.
\newblock In {\em Proceedings of the IEEE/CVF Conference on Computer Vision and
  Pattern Recognition}, pages 5376--5385, 2020.

\bibitem{3dssl-partseg}
Jonathan Sauder and Bjarne Sievers.
\newblock Self-supervised deep learning on point clouds by reconstructing
  space.
\newblock {\em Advances in Neural Information Processing Systems}, 32, 2019.

\bibitem{waymo}
Pei Sun, Henrik Kretzschmar, Xerxes Dotiwalla, Aurelien Chouard, Vijaysai
  Patnaik, Paul Tsui, James Guo, Yin Zhou, Yuning Chai, Benjamin Caine, et~al.
\newblock Scalability in perception for autonomous driving: Waymo open dataset.
\newblock In {\em Proceedings of the IEEE/CVF conference on computer vision and
  pattern recognition}, pages 2446--2454, 2020.

\bibitem{taubin}
Gabriel Taubin.
\newblock Estimating the tensor of curvature of a surface from a polyhedral
  approximation.
\newblock In {\em Proceedings of IEEE International Conference on Computer
  Vision}, pages 902--907. IEEE, 1995.

\bibitem{kpconv}
Hugues Thomas, Charles~R Qi, Jean-Emmanuel Deschaud, Beatriz Marcotegui,
  Fran{\c{c}}ois Goulette, and Leonidas~J Guibas.
\newblock Kpconv: Flexible and deformable convolution for point clouds.
\newblock In {\em Proceedings of the IEEE/CVF international conference on
  computer vision}, pages 6411--6420, 2019.

\bibitem{occo}
Hanchen Wang, Qi Liu, Xiangyu Yue, Joan Lasenby, and Matt~J Kusner.
\newblock Unsupervised point cloud pre-training via occlusion completion.
\newblock In {\em Proceedings of the IEEE/CVF international conference on
  computer vision}, pages 9782--9792, 2021.

\bibitem{dgcnn}
Yue Wang, Yongbin Sun, Ziwei Liu, Sanjay~E Sarma, Michael~M Bronstein, and
  Justin~M Solomon.
\newblock Dynamic graph cnn for learning on point clouds.
\newblock {\em Acm Transactions On Graphics (tog)}, 38(5):1--12, 2019.

\bibitem{maskfeat}
Chen Wei, Haoqi Fan, Saining Xie, Chao-Yuan Wu, Alan Yuille, and Christoph
  Feichtenhofer.
\newblock Masked feature prediction for self-supervised visual pre-training.
\newblock In {\em Proceedings of the IEEE/CVF Conference on Computer Vision and
  Pattern Recognition}, pages 14668--14678, 2022.

\bibitem{welch1994free}
William Welch and Andrew Witkin.
\newblock Free-form shape design using triangulated surfaces.
\newblock In {\em Proceedings of the 21st annual conference on Computer
  graphics and interactive techniques}, pages 247--256, 1994.

\bibitem{pointcontrast}
Saining Xie, Jiatao Gu, Demi Guo, Charles~R Qi, Leonidas Guibas, and Or Litany.
\newblock Pointcontrast: Unsupervised pre-training for 3d point cloud
  understanding.
\newblock In {\em European conference on computer vision}, pages 574--591.
  Springer, 2020.

\bibitem{simmim}
Zhenda Xie, Zheng Zhang, Yue Cao, Yutong Lin, Jianmin Bao, Zhuliang Yao, Qi
  Dai, and Han Hu.
\newblock Simmim: A simple framework for masked image modeling.
\newblock In {\em Proceedings of the IEEE/CVF Conference on Computer Vision and
  Pattern Recognition}, pages 9653--9663, 2022.

\bibitem{second}
Yan Yan, Yuxing Mao, and Bo Li.
\newblock Second: Sparsely embedded convolutional detection.
\newblock {\em Sensors}, 18(10):3337, 2018.

\bibitem{3dssd}
Zetong Yang, Yanan Sun, Shu Liu, and Jiaya Jia.
\newblock 3dssd: Point-based 3d single stage object detector.
\newblock In {\em Proceedings of the IEEE/CVF conference on computer vision and
  pattern recognition}, pages 11040--11048, 2020.

\bibitem{centerpoint}
Tianwei Yin, Xingyi Zhou, and Philipp Krahenbuhl.
\newblock Center-based 3d object detection and tracking.
\newblock In {\em Proceedings of the IEEE/CVF conference on computer vision and
  pattern recognition}, pages 11784--11793, 2021.

\bibitem{Point-bert}
Xumin Yu, Lulu Tang, Yongming Rao, Tiejun Huang, Jie Zhou, and Jiwen Lu.
\newblock Point-bert: Pre-training 3d point cloud transformers with masked
  point modeling.
\newblock In {\em Proceedings of the IEEE/CVF Conference on Computer Vision and
  Pattern Recognition}, pages 19313--19322, 2022.

\bibitem{zhang2008curvature}
Xiaopeng Zhang, Hongjun Li, Zhanglin Cheng, et~al.
\newblock Curvature estimation of 3d point cloud surfaces through the fitting
  of normal section curvatures.
\newblock {\em Proceedings of ASIAGRAPH}, 2008:23--26, 2008.

\bibitem{depthcontrast}
Zaiwei Zhang, Rohit Girdhar, Armand Joulin, and Ishan Misra.
\newblock Self-supervised pretraining of 3d features on any point-cloud.
\newblock In {\em Proceedings of the IEEE/CVF International Conference on
  Computer Vision}, pages 10252--10263, 2021.

\bibitem{ibot}
Jinghao Zhou, Chen Wei, Huiyu Wang, Wei Shen, Cihang Xie, Alan Yuille, and Tao
  Kong.
\newblock ibot: Image bert pre-training with online tokenizer.
\newblock {\em arXiv preprint arXiv:2111.07832}, 2021.

\bibitem{dynamic-voxelization}
Yin Zhou, Pei Sun, Yu Zhang, Dragomir Anguelov, Jiyang Gao, Tom Ouyang, James
  Guo, Jiquan Ngiam, and Vijay Vasudevan.
\newblock End-to-end multi-view fusion for 3d object detection in lidar point
  clouds.
\newblock In {\em Conference on Robot Learning}, pages 923--932. PMLR, 2020.

\bibitem{voxelnet}
Yin Zhou and Oncel Tuzel.
\newblock Voxelnet: End-to-end learning for point cloud based 3d object
  detection.
\newblock In {\em Proceedings of the IEEE conference on computer vision and
  pattern recognition}, pages 4490--4499, 2018.

\bibitem{cbgs}
Benjin Zhu, Zhengkai Jiang, Xiangxin Zhou, Zeming Li, and Gang Yu.
\newblock Class-balanced grouping and sampling for point cloud 3d object
  detection.
\newblock {\em arXiv preprint arXiv:1908.09492}, 2019.

\bibitem{cylinder3d}
Xinge Zhu, Hui Zhou, Tai Wang, Fangzhou Hong, Yuexin Ma, Wei Li, Hongsheng Li,
  and Dahua Lin.
\newblock Cylindrical and asymmetrical 3d convolution networks for lidar
  segmentation.
\newblock In {\em Proceedings of the IEEE/CVF conference on computer vision and
  pattern recognition}, pages 9939--9948, 2021.

\bibitem{pointshop}
Matthias Zwicker, Mark Pauly, Oliver Knoll, and Markus Gross.
\newblock Pointshop 3d: An interactive system for point-based surface editing.
\newblock {\em ACM Transactions on Graphics (TOG)}, 21(3):322--329, 2002.

\end{thebibliography}
}

\end{document}